\documentclass[letterpaper, 10 pt, conference]{ieeeconf}  
\let\labelindent\relax
\usepackage[utf8]{inputenc} %
\usepackage[T1]{fontenc}    %
\usepackage{graphicx} \graphicspath{ {figures/} }
\usepackage{amsmath,amssymb,mathabx,amsbsy,mathtools,etoolbox}
\usepackage{times}
\usepackage{graphicx}
\usepackage{amsmath,amssymb}
\usepackage{acronym}
\usepackage{enumitem}
\usepackage[breaklinks,colorlinks,linkcolor=black]{hyperref}
\usepackage{balance}
\usepackage{xspace}
\usepackage{setspace}
\usepackage[skip=3pt,font=small]{subcaption}
\usepackage[skip=3pt,font=small]{caption}
\usepackage[dvipsnames,svgnames,x11names]{xcolor}
\usepackage[capitalise,noabbrev,nameinlink]{cleveref}
\usepackage{booktabs,tabularx,colortbl,multirow,multicol,array,makecell}
\usepackage{dblfloatfix}
\usepackage[misc]{ifsym}
\usepackage{pifont}
\usepackage{cite}
\usepackage{graphicx}
\usepackage{tikz}
\newcommand{\circled}[1]{\tikz[baseline=(char.base)]{\node[shape=circle,draw,inner sep=0.5pt] (char) {#1};}}

\makeatletter
\DeclareRobustCommand\onedot{\futurelet\@let@token\@onedot}
\def\@onedot{\ifx\@let@token.\else.\null\fi\xspace}

\def\etal{\emph{et al}\onedot}
\makeatother

\DeclareMathOperator*{\argmin}{argmin}

\makeatletter
\def\BState{\State\hskip-\ALG@thistlm}
\makeatother

\makeatletter
\renewcommand{\paragraph}{%
  \@startsection{paragraph}{4}%
  {\z@}{0ex \@plus 0ex \@minus 0ex}{-1em}%
  {\hskip\parindent\normalfont\normalsize\bfseries}%
}
\makeatother

\crefname{equation}{Eq.\!\!}{equations}
\crefname{algorithm}{Alg.}{Algs.}
\Crefname{algocf}{Algorithm}{Algorithms}
\crefname{section}{Sec.}{Secs.}
\Crefname{section}{Section}{Sections}
\crefname{figure}{Fig.\!\!}{Fig.}
\Crefname{figure}{Figure}{Figure}

\definecolor{gblue}{HTML}{4285F4}
\definecolor{gred}{HTML}{DB4437}
\definecolor{ggreen}{HTML}{0F9D58}

\definecolor{mygray}{gray}{.92}

\usepackage{pifont}
\usepackage{bbm}
\usepackage{soul}
\usepackage[export]{adjustbox}

\IEEEoverridecommandlockouts                              

\def\etal{\mbox{\textit{et al.}}\@\xspace}

\title{\centering \LARGE \bf Toward Spatial-Temporal Consistency of Joint Visual-Tactile Perception in VR Applications}

\author{Fuqiang Zhao$^{1}$, Kehan Zhang$^{1}$, Qian Liu$^{1 \ast}$, Zhuoyi Lyu$^{2}$
\thanks{This work was supported in part by the National Science Foundation of China (Grant No.62071083), and in part by the Dalian Science and Technology Innovation Foundation (No. 2022JJ12GX014).}
\thanks{${^1}$These authors are with the Department of Computer Science and Technology, Dalian University of Technology, China. Emails: fuqiangzh@mail.dlut.edu.cn, kehan\_zhang@mail.dlut.edu.cn, qianliu@dlut.edu.cn.
}
\thanks{${^2}$This author is with the vivo Mobile Communication Co., Ltd, China. Emails: zhuoyi.lv@vivo.com
}
\thanks{
The first two authors contribute equally.
}
\thanks{$^{\ast}$Corresponding author: Qian Liu.}
}

\begin{document}

\maketitle
\thispagestyle{empty}
\pagestyle{empty}

\begin{abstract}

With the development of VR technology, especially the emergence of the metaverse concept, the integration of visual and tactile perception has become an expected experience in human-machine interaction. Therefore, achieving spatial-temporal consistency of visual and tactile information in VR applications has become a necessary factor for realizing this experience. The state-of-the-art vibrotactile datasets generally contain temporal-level vibrotactile information collected by randomly sliding on the surface of an object, along with the corresponding image of the material/texture. However, they lack the position/spatial information that corresponds to the signal acquisition, making it difficult to achieve spatiotemporal alignment of visual-tactile data. Therefore, we develop a new data acquisition system in this paper which can collect visual and vibrotactile signals of different textures/materials with spatial and temporal consistency. In addition, we develop a VR-based application call “V-Touching” by leveraging the dataset generated by the new acquisition system, which can provide pixel-to-taxel joint visual-tactile perception when sliding over the surface of objects in the virtual environment with distinct vibrotactile feedback of different textures/materials. Our data and code are available at \url{https://github.com/wmtlab/Pixel2Taxel}

\end{abstract}

\section{Introduction} \label{Intro}

Tactile perception, as one of the most important human senses, plays a significant role in direct interaction and perception of the physical environment. In recent years, various tactile technologies have been widely applied in the realm of virtual reality (VR)~\cite{steinbach2018haptic,song2016finding,tong2023survey} to enhance the immersive perception of users. A typical example is to provide users with vibrotactile sensations when touching a virtual object and “feeling” the texture and shape of it.

Despite significant advancements in tactile-related technologies, it remains a major challenge to accurately simulate the sense of touch when interacting with virtual objects. This problem becomes even more crucial when the texture features of virtual objects are heterogeneous, where different regions of the texture exhibit distinct characteristics. For example, it is common in real life to see an object with a smooth and patternless surface in one part, while another part has sophisticated patterns. Existing work~\cite{5963667,6954342,8816167, 6775475,8460494} can only describe the overall texture in relation to certain tactile signals, but cannot achieve a pixel-to-taxel mapping between the texture image and the corresponding vibrotactile signal due to the lack of position/spatial information during the data acquisition process.

\begin{figure}[t!]
\centering
\vspace{0.25cm}
\includegraphics[width=0.48\textwidth]{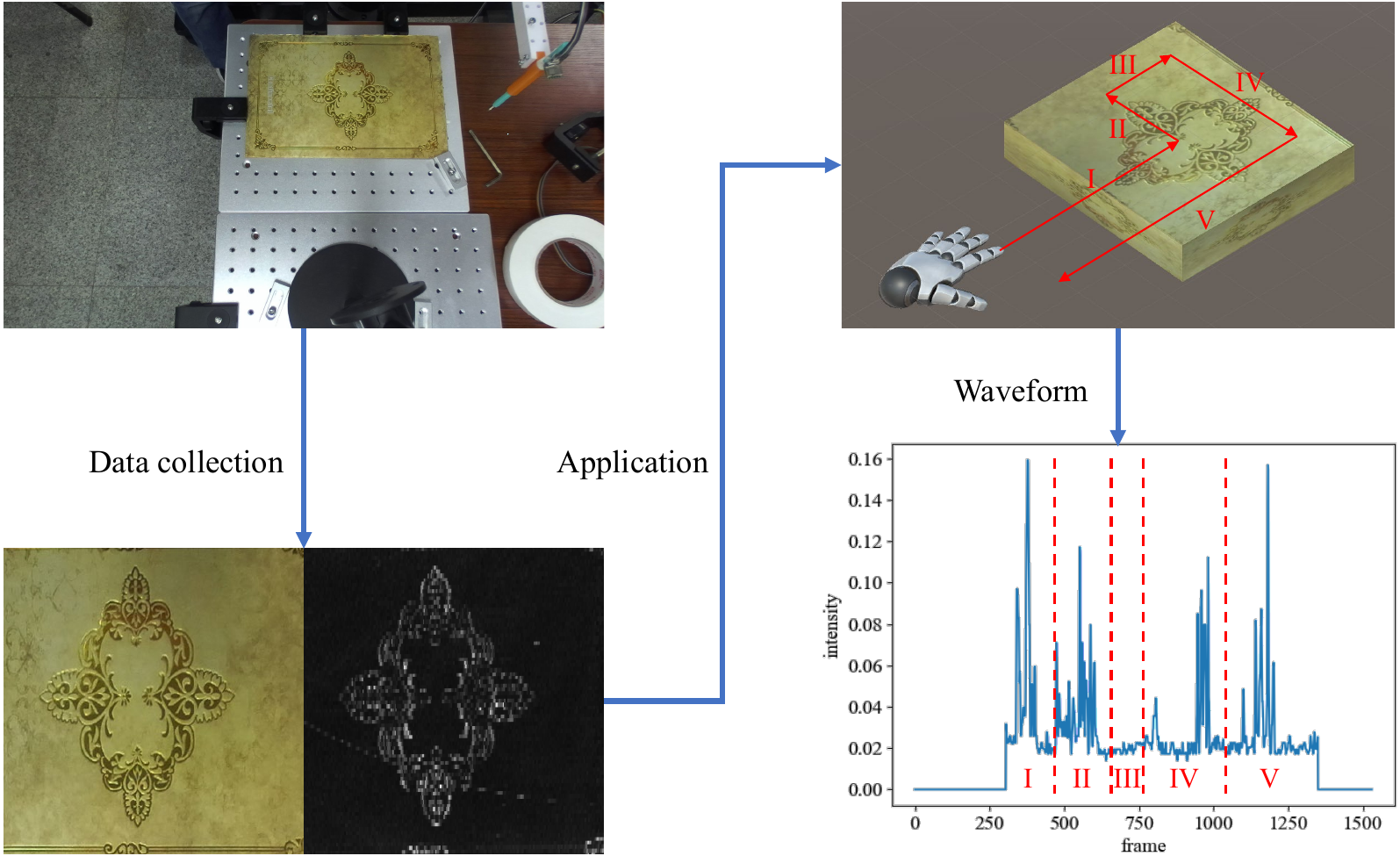}
\caption{\textbf{The pipeline of our work.} We first collect vibrotactile data with an accelerometer by sliding back and forth on the surface of physical objects. These data are then used to establish a vibration map with spatial alignment toward the captured texture images. We then develop a VR application, called \textbf{V-Touching}, by utilizing the dataset generated by the proposed acquisition system. The user controls a haptic glove to interact with the surface of virtual objects, and the server transmits corresponding vibrotactile signals to the client with satisfactory spatial-temporal alignment of visual-tactile perception.}
\label{fig:pipline}
\end{figure}

The pipeline of our work shown in ~\cref{fig:pipline} consists of two stages: 

\textbf{1)} We propose a new data collection system, which scans the visual and tactile information across various textures simultaneously and establish a spatiotemporal mapping between visual and tactile signals based on coordinates. The developed scheme enables the creation of a spatially consistent visual-tactile datasets. In this paper, we adopt the acceleration information to mimic the tactile signal when contacting or sliding over the surface of an object as that of~\cite{8547512,6954342}. Then, we utilize the collected vibrotactile signal to build a vibration map with taxel-level consistency towards the corresponding texture image.

\textbf{2)} In order to validate the generated dataset, we develop an interactive VR application, called \textbf{V-Touching}. This system allows users to experience realistic vibrotactile sensations while using virtual hands to explore the textures of objects at different spatial positions. In contrast to existing texture related VR applications, this game can provide spatiotemporal aligned experience for visual and tactile sensation. In particular, the user can feel expected/realistic vibrotactile feedback when touching surfaces with either homogeneous or  heterogeneous patterns in the virtual environment.

\section{Related Work}\label{RW}

Many work investigated the relationship between the tactile information and surface properties of objects~\cite{WHC15,Culbertson,Sinapov,richardson2019improving,9765606}, typically involving direct or indirect analysis of tactile signals to classify objects based on their features. The inputs to these approaches usually consisted of a series of vibrotactile signals collected by an accelerometer via sliding the surface of objects, with the corresponding object categories as the output. Although some studies~\cite{gao2016deep,TUM,strese2019haptic} used both visual and tactile features to predict object categories, but cannot demonstrate satisfactory performance on objects with complex surface textures. This was caused by the limitations of existing datasets adopted in these research. 

\begin{figure}[t!]
\centering
\includegraphics[width=0.49\textwidth]{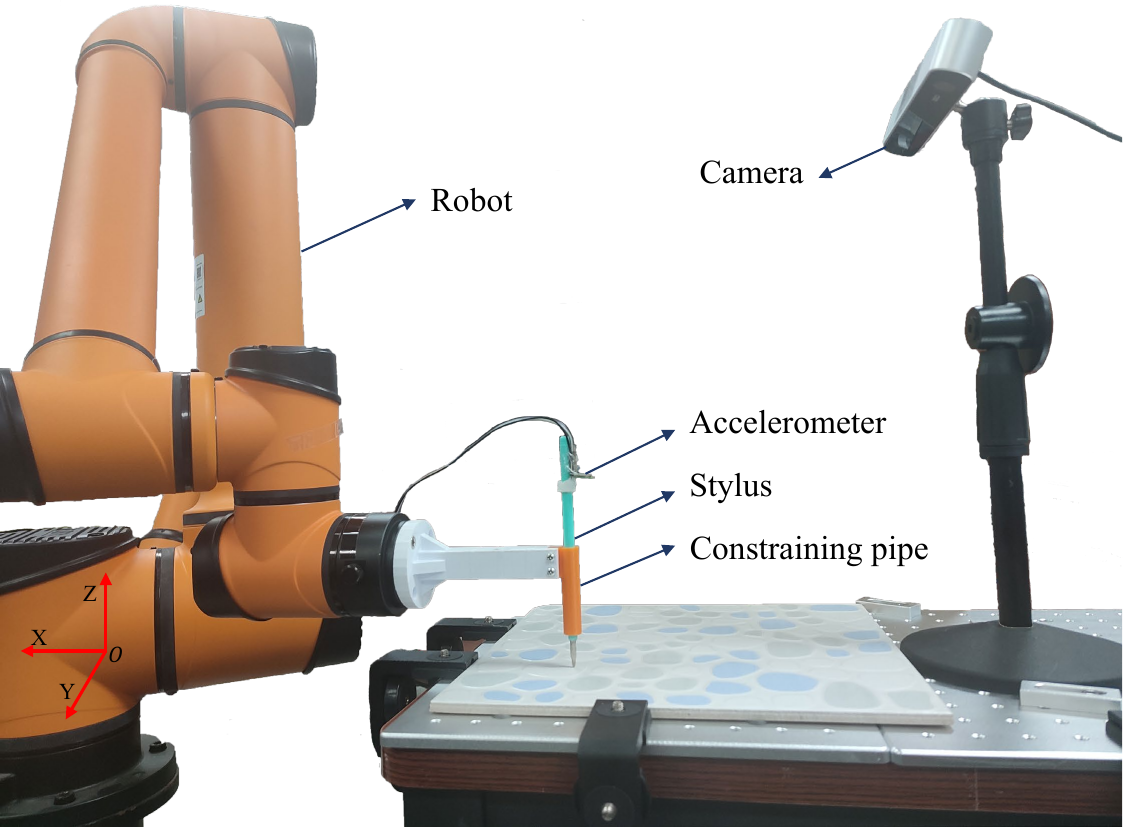}
\caption{\textbf{System Setups.} The developed system consists of both visual and tactile data acquisition devices. With the origin of the world coordinate mounted on the base of the robotic arm, as shown by the red coordinate system in the bottom left corner.}
\label{fig:setup}
\end{figure}

In \cite{6954342}, the popular LMT haptic texture dataset was generated by using an accelerometer mounted on a stylus sliding across various textured objects. However, due to the arbitrary nature of the sliding trajectory, this dataset was primarily suitable for texture classification tasks but cannot be effectively utilized to construct an application with joint visual-tactile sensations that strongly correlates with spatial positions. Similarly, in the work presented in \cite{5963667}, the authors collected stylus coordinates during the acquisition of acceleration data. However, these coordinates were only utilized to calculate the horizontal velocity of the stylus. The calibration of vertical acceleration and horizontal coordinates was not performed. In addition, the texture data collection process was carried out manually, without a predetermined trajectory, which made it challenging to cover the entire sampling area. In \cite{8816167}, the tactile information was collected with human finger touch fabrics. The motion trajectory of the finger was constrained with tracks. A grating ruler sensor was employed to measure the displacement (coordinates). However, the data collection was limited to a straight line, resulting in the acquisition of one-dimensional coordinate data instead of two-dimensional spatial information.

In summary, the existing data acquisition system and the corresponding generated datasets cannot provide effective spatial-temporal consistency of visual-tactile signals. This is the reason why we develop a new data collection system in this paper beneficial for applications with joint visual and tactile experience.

   
\begin{figure}[t!]
\begin{subfigure}{0.49\linewidth}
    \centering
    \includegraphics[width=\linewidth]{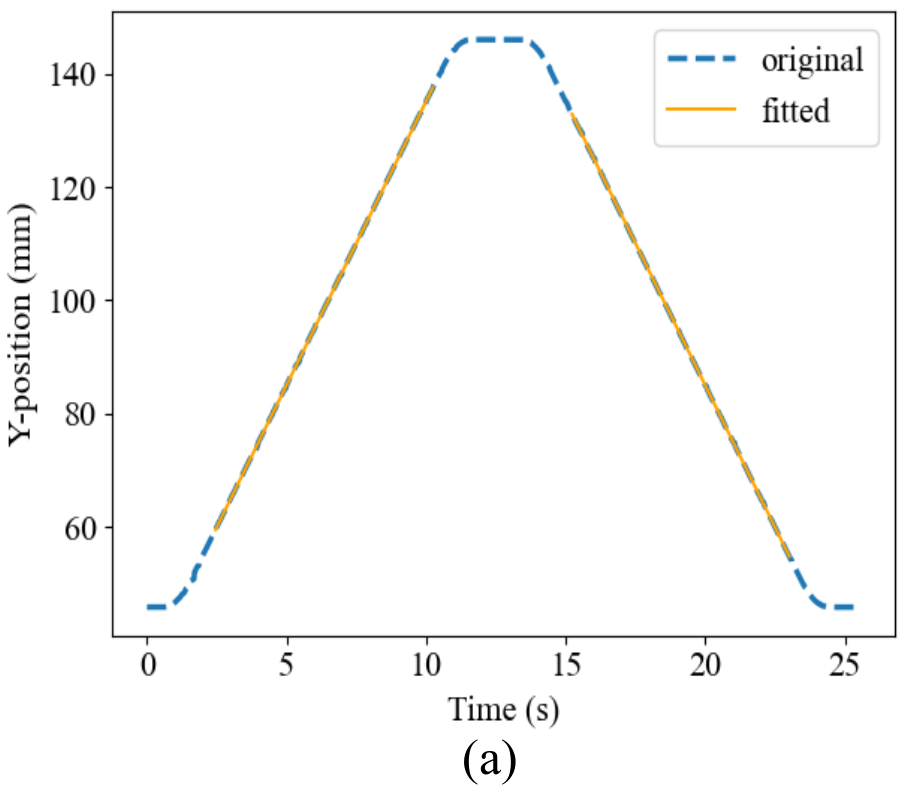}
    \phantomsubcaption
    \label{fig:Y_position}
\end{subfigure}
\begin{subfigure}{0.49\linewidth}
    \centering
    \includegraphics[width=\linewidth]{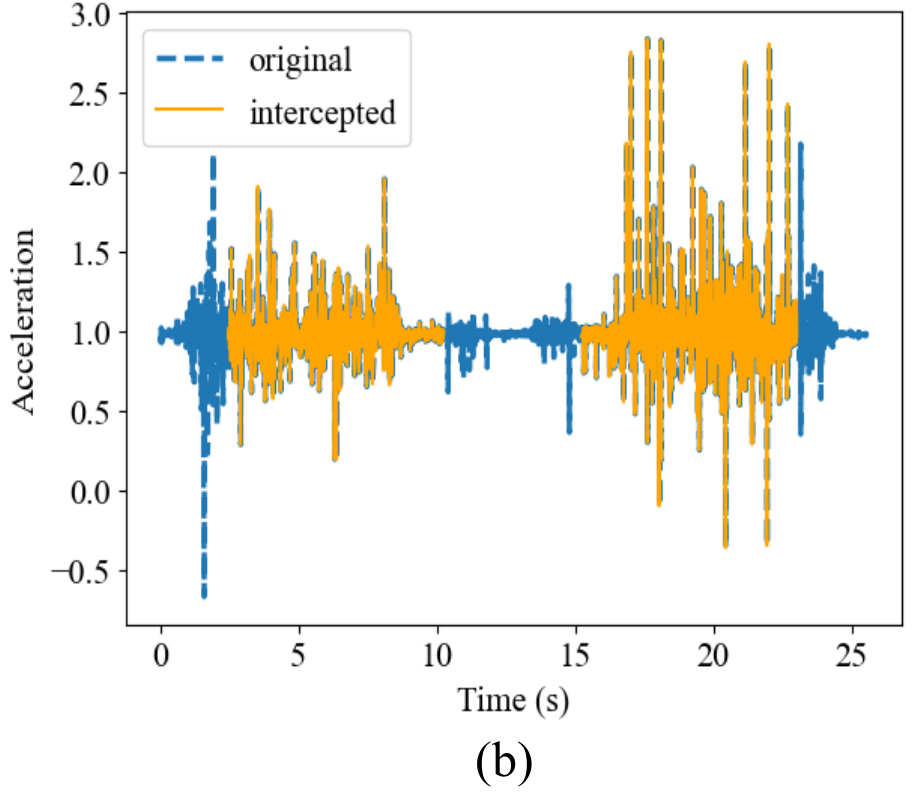}
    \phantomsubcaption
    \label{fig:Acc}
\end{subfigure}
\vspace{-0.4cm} 
\caption{\textbf{a)}: The blue dashed line represents the position variation signal acquired by the robot, while the orange line represents the fitted signal of the robot's uniform motion. \textbf{b)}: The blue dashed line represents the raw signal collected  by the accelerometer, while the orange line represents the intercepted accelerometer signal.}
\end{figure}

\begin{figure*}[t!]
\centering
\includegraphics[width=0.9\textwidth,height=0.5\textwidth]{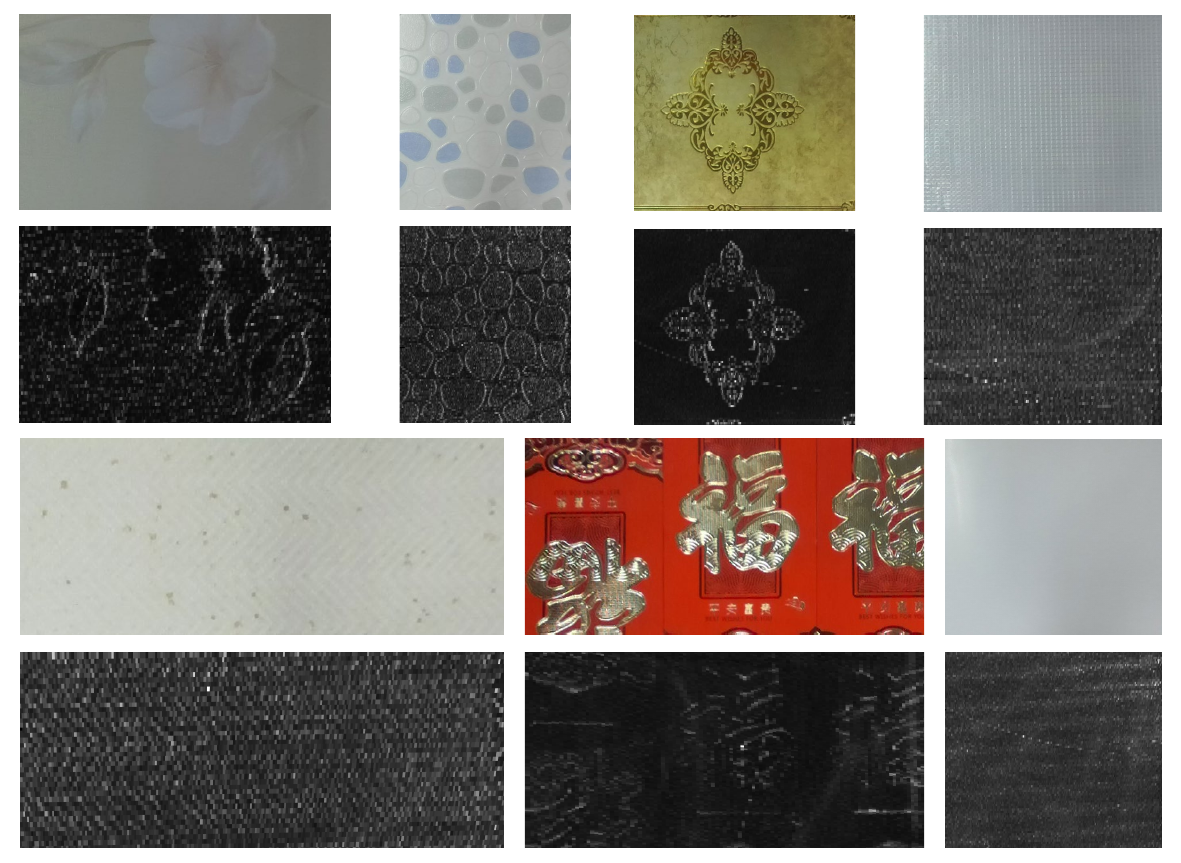}
\caption{\textbf{Qualitative results of data collection.} The \textbf{\textit{top}} shows the captured texture images, and the \textbf{\textit{bottom}} displays the generated vibration maps of collected vibrotactile data, where brighter areas indicate stronger intensity.}
\label{fig:result}
\end{figure*}

\section{A New Data Acquisition System With Spatiotemporal Consistency}\label{Method}

In order to collect spatiotemporal consistent visual-tactile data, two mappings are essential: \textbf{(1)}\;the mapping between the position of the stylus tip in the world coordinate system and its corresponding position in the captured texture image, and \textbf{(2)}\;the mapping between the position of the stylus tip in the world coordinate system and the corresponding vibrotactile data collected by the accelerometer. In this section, we will introduce our experimental setups and how to realize these two mappings in details.

\subsection{System Setups}
The data acquisition system is shown in ~\cref{fig:setup}, consisting of both visual and tactile capture devices. We use a camera fixed on a metal bracket to capture the original images of the visual texture information. The tactile capture device includes a robotic arm, a stylus, a constraining pipe, and an accelerometer. The robotic arm is used to control the movement of the stylus, which is equipped with an accelerometer (WT9011G4K from Witmotion Co. in this paper) and contacts with the surface of different objects with homogeneous and heterogeneous patterns. In this research, the acceleration data in the vertical direction (i.e. Z-axis) is particularly important. Therefore, we use a 3D-printed constraining pipe to the stylus. Specifically, the stylus has a diameter of 0.80cm, and the internal diameter of the constraining pipe is 0.85cm. This pipe is designed precisely to limit the horizontal movement of the stylus while imposing minimal friction that could hinder its vertical movement. It also ensures that the stylus remains in a straight line along the sliding direction. The stylus consists of a metal cone-shaped tip and a stylus shaft. The metal cone-shaped tip enhances its sensitivity during the vibrotactile signal collection process. As the stylus slides across the surface of the object, it captures the vertical vibrotactile information corresponding to the surface texture. We consider the acceleration signals captured by the accelerometer as the vibrotactile signals in this paper.

During the tactile information collection process, we first secure that the object is fixed onto a plate without any movements. Then, we control the robotic arm to move along the Y-axis. For different objects, we adjust the maximum distance that the stylus can travel to ensure it always within the boundaries of the object. When the stylus reaches the maximum distance, it reverses its direction and continues scanning. Specifically, we set the stylus to traverse the straight-line region four times, both in the positive and negative directions along the Y-axis. After completing one trial, the robotic arm moves 2mm in the positive X-direction, then continues to capture the vibrotactile data when moving along the Y-direction. The robotic arm repeats this process and stops at the predetermined maximum boundary.

\subsection{Pixel-to-Taxel Mapping}    

\textbf{Mapping from World to Pixel:} The mapping from the world position of the tip to the pixel position can be obtained by applying camera calibration methods as described by Zhang \etal~in \cite{zhang2000flexible}. Specifically, we first mark a certain number $(n \geq 8)$ of points in dispersed areas on the object's surface, then sequentially move the robotic arm to these marked points and record their corresponding world coordinates $P = \left(X, Y, Z\right)$. The world coordinates here refer to the coordinates of the end effector in the base coordinate system of the robotic arm. Since the objects chosen in our experiment can be approximated as planar objects, we fix the $Z$ coordinate of the object in the world coordinate system to 0. We also mark the corresponding pixel coordinates $p = \left(x, y\right)$ in the captured images of the object. Therefore, we have a set of mapped points between the object's world coordinates and the pixel coordinates.

We then use the camera calibration algorithm with the obtained set of corresponding points. This allows us to obtain the camera's intrinsic matrix $M_{\text{in}} \in \mathbb{R}^{3 \times 3}$, and extrinsic matrix $M_{\text{ex}} \in \mathbb{R}^{3 \times 4}$. The projection relationship can be expressed as follows:
\begin{equation}
\begin{bmatrix}x \\y \\1 \\ \end{bmatrix} 
\ =\  \frac{1}{C} M_{\text{in}} \ast M_{\text{ex}} \ast 
\begin{bmatrix}X \\Y \\Z \\1 \\\end{bmatrix}  
\end{equation}
where the symbol \textbf{$``\ast"$} represents the matrix multiplication. $C$ represents the scale factor, which is the effective focal length and indicates the distance between the object and the optical center. This implies that during the camera calibration, if an object is at different distances from the camera, we need to calibrate the camera at different positions. Therefore, considering the data collection process, given objects with different heights, their relative positions with respect to the camera will vary, leading to significant errors in the mapping relationship. Therefore, whenever we change to an object with a different height, we need to recalibrate the camera. To simplify the expression, we denote the projection as $f$. Then, we have $p = f(P)$. Through $f$, we can project any point in the world coordinate system onto the pixel coordinate system.

\begin{figure*}[t!]
\vspace{0.1cm}
\includegraphics[width=\textwidth,height=0.35\textwidth]{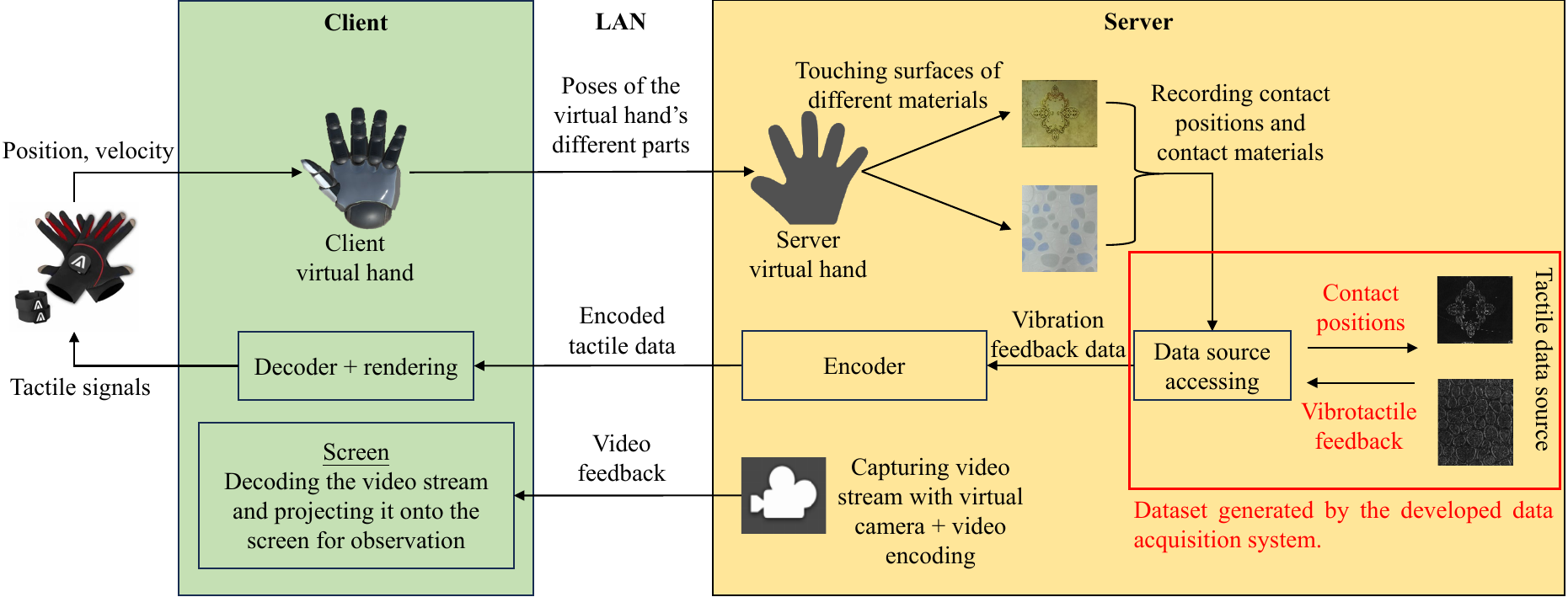}
\caption{\textbf{An illustration of the framework of the developed VR application, \textbf{V-Touching}.} In this application, the client is able to control the poses of different parts of the hand. The server executes the movement of these poses to touch virtually textured objects. It encodes the vibrotactile signals and transmits them to the client. The client then decodes the signals, resulting in the haptic glove generating vibrations corresponding to specific positions.}
\label{fig:application}
\end{figure*}

\textbf{Mapping from Taxel to World:} Although the relative positions of the stylus and the accelerometer in the horizontal direction remain consistent, there is a mismatch between the sampling rate of the accelerometer and the sampling position of the robotic arm's end effector. Data processing is needed to align the two sets of data. Specifically, the data sequences obtained from the robotic arm are denoted as $\mathbf{R} = \{(t_i, X, Y_i)\}^N_{i=1}$, and the data sequences obtained from the accelerometer are denoted as $\mathbf{A} = \{(t_j, Acc_j)\}^M_{j=1}$, where $N$ and $M$ represent the lengths of the two sampling sequences, $t_i$ and $t_j$ denote the timestamps of the data samples, $X$ and $Y_i$ represent the positions sampled by the robotic arm in the X and Y directions, respectively. For each sequence, the position of the robotic arm in the X-axis remains unchanged, indicating that $X$ is a constant. $Acc_j$ represents the vertically sampled acceleration from the accelerometer. It is noted that we only illustrate the data needed in the Taxel-to-World mapping here, rather than all the data obtained during the acquisition process.

The blue dashed line in~\cref{fig:Y_position} illustrates the variation of the Y-axis position acquired by the robotic arm during its back-and-forth motion. We can be observe that the robotic arm undergoes acceleration initially, followed by a period of constant velocity, and finally deceleration. Due to the unstable position changes during the acceleration and the deceleration, we only utilize the data acquired during the constant velocity phase of the robotic arm's motion. The start and end indices of the sequence corresponding to the constant velocity phase can be obtained using the~\cref{i_1} and~\cref{i_2}, respectively.
\begin{equation} \label{i_1}
    i_1\ =\ \argmin_i\left(\left|Y_i-(\bar{Y}-w l)\right|\right)
\end{equation}
\begin{equation} \label{i_2}
    i_2\ =\ \argmin_i\left(\left|Y_i-(\bar{Y}+w l)\right|\right)
\end{equation}
where the symbol $``|\ |"$ represents the absolute operation. $\bar{Y}$ represents the midpoint position of the total movement range of the robotic arm. It can be approximated by the average value of $\bar{Y} = \left(\frac{1}{N}\sum_{i}^{N}Y_i\right)$. $l$ represents the total length of the robotic arm's movement along the Y direction, and $w$ is a weight parameter with a range of $0 < w \leq 0.5$. A smaller value of $w$ increases the probability of obtaining indices within the constant velocity phase, but it also shortens the resulting motion interval. In this paper, we set $w=0.45$. Since the robotic arm performs back-and-forth motion along the Y-axis, when it moves in the positive Y direction, the start index and end index are denoted as $i_1$ and $i_2$, respectively. Conversely, when the robotic arm moves in the negative Y direction, the start index and end index are reversed, denoted as $i_2$ and $i_1$, respectively.

Taking the example of the robotic arm moving in the positive Y direction, once we obtain the indices $i_1$ and $i_2$, we can directly determine the start and the end time of the constant velocity phase as $t_{i_1}$ and $t_{i_2}$, respectively. Furthermore, we can utilize the least square method to fit the motion equation of the robotic arm during the constant velocity phase, which can be expressed as
\begin{equation}\label{Yt}
Y = mt + b \hspace{0.5cm}t \in \left(t_{i_1},t_{i_2}\right)
\end{equation}
where $m$ represents the slope and $b$ represents the Y-intercept. By using this equation, we can input any time $t$ within an appropriate range to obtain the position of the robotic arm at that specific moment. The fitted position variation, depicted by the orange line in~\cref{fig:Y_position}, closely matches the actual position variation. The orange line in ~\cref{fig:Acc} represents the acceleration signal that we intercepted. It is more accurate because it is not affected by the original signal collected during acceleration and deceleration phases.

For the timestamp indices corresponding to the acceleration sampling sequence $\mathbf{A}$, we have
\begin{equation}\label{j_1}
j_1 \ =\ \argmin_j\left(\left|t_j - t_{i_1}\right|\right) + 1
\end{equation}
\begin{equation}\label{j_2}
j_2 \ =\ \argmin_j\left(\left|t_j - t_{i_2}\right|\right) - 1
\end{equation}

Therefore, for each acceleration $Acc_j$ and its corresponding timestamp $t_j \in \left[t_{j_1}, t_{j_2}\right]$, we can utilize the~\cref{Yt} to obtain the corresponding position $P_j$ of the robotic arm. We simplify this mapping process as $g$, represented by $P=g(Acc,t)$, where $t \in \left[t_{j_1}, t_{j_2}\right]$.

The robotic arm's data collection process involves acquiring 8 sets of data along a straight-line trajectory, each containing $K_i$ samples. Hence, utilizing the formula $g$, we can establish a data sequence $\mathbf{D}=\{(Acc_k,P_k)\}_{k=1}^{\sum_{i=1}^{8}{\!K_i}}$, where each vibrotactile information is paired with its corresponding position.

Furthermore, we should point out that: \textbf{(1)}\;the robotic arm collects data in a back-and-forth manner, resulting in unordered positions in the $\mathbf{D}$ sequence. To address this issue, we sort the sequence $\mathbf{D}$ based on positions, ensuring that the positions in $\mathbf{D}$ are in the ascending order; \textbf{(2)}\;for locations on the surface of object that have protrusions, the directions of the accelerations generated while the robotic arm's movement are opposite. To account for this, we transform the accelerations as $V = \left|Acc - 1\right|$, representing the vibration intensity, where the $``1"$ represents the baseline acceleration, which is typically the gravitational acceleration.

Analogy to pixels and resolution in images, we have also define the resolution of the vibration maps with taxels in this research. Specifically, we take the average of all vibration intensity within a 1mm interval to obtain the descriptive value for each taxel. This approach helps to reduce the impact of outliers during the data collection process. Finally, we project the taxels onto pixels by the formula $g$ and $f$. Since the interval of a taxel is 1mm, it projects to multiple pixels along a straight line. Therefore, we stretch them by 3 pixel units on both sides. This means that one taxel corresponds to a small area of pixels in the texture image. In this paper, the image resolution is 1920×1080, so this stretching operation does not significantly impact the our final results. The stretching ensures that the information of each taxel is distributed across a small region in the image, maintaining a reasonable representation within the image resolution. 

In addition, we apply the min-max normalization to the constructed vibration maps, which follows
\begin{equation}
V' \ =\ \frac{V- V_{\text{min}}}{V_{\text{max}} - V_\text{{min}}}
\end{equation}
After normalization, we can convert the vibration map into image format for file storage. Since our camera is not placed horizontally, we need to perform perspective transformation on the image and vibration maps to eliminate distortion.

\subsection{Results and Analysis}
The results are shown in ~\cref{fig:result}, where we present the captured results of seven objects. In the first row, from left to right, there are the following items: a pattern board, a relief ceramic tile, a notebook cover, and a plastic grid. Moving on to the second row, there are the following items: a gift box, a red envelope, and a plastic cutting board. The gift box is a decorative container used for presenting gifts on special occasions It is important to note that some of our captured results exhibit curved lines. This is because these objects are not rigid and inherently possess some degree of bending. This explains why these curves appear so smooth. Comparing the ceramic tiles with the other objects, it is evident that the tiles do not exhibit this issue, indicating that better results can be obtained when capturing the texture of rigid objects.
 
In~\cref{table:result}, we present the quantitative results of each objective in the same order of textures presented in~\cref{fig:result}, including the vibration amplitude $V_{\text{scale}} = V_{\text{max}} - V_{\text{min}}$, the average vibration intensity $V_{\text{mean}}$, and the standard deviation of the vibration maps $V_{\text{std}}$. The quantitative results increase as the object surfaces become harder and rougher. Moreover, the surface of red envelope characterized by prominent edge variations but an overall flatness, results in a larger value for $V_{\text{scale}}$. These characteristics confirm that the developed scheme can yield reliable and accurate data acquisition.

\begin{table}[tb!]    
\centering
\caption{\label{table:result}  Quantitative Results} 
\begin{tabular}{@{\hspace{8pt}}c@{\hspace{10pt}}c@{\hspace{8pt}}c@{\hspace{8pt}}c@{\hspace{8pt}}c@{\hspace{8pt}}c@{\hspace{8pt}}c@{\hspace{8pt}}c@{\hspace{8pt}}}
  \hline
  \rule{0pt}{10pt}
  \ & \circled{1} & \circled{2} & \circled{3} & \circled{4} & \circled{5} & \circled{6} & \circled{7} \\
   $V_{\text{scale}}$ &\small0.666 & \small0.966 & \small0.254 &\small0.128 & \small0.198 & \small0.269 &\small0.103  \\
   $V_{\text{mean}}$  &\small0.065 & \small0.173 & \small0.026 &\small0.029 & \small0.045 & \small0.030 &\small0.024   \\
   $V_{\text{std}}$ & \small0.052 &\small0.086 &\small0.019 &\small0.008 &\small0.018 &\small0.014 &\small0.005 \\
  \hline
\end{tabular}
\end{table}

\begin{figure*}[t!]
\centering
\vspace{0.1cm}
\includegraphics[width=0.95\textwidth,height=0.321\textwidth]{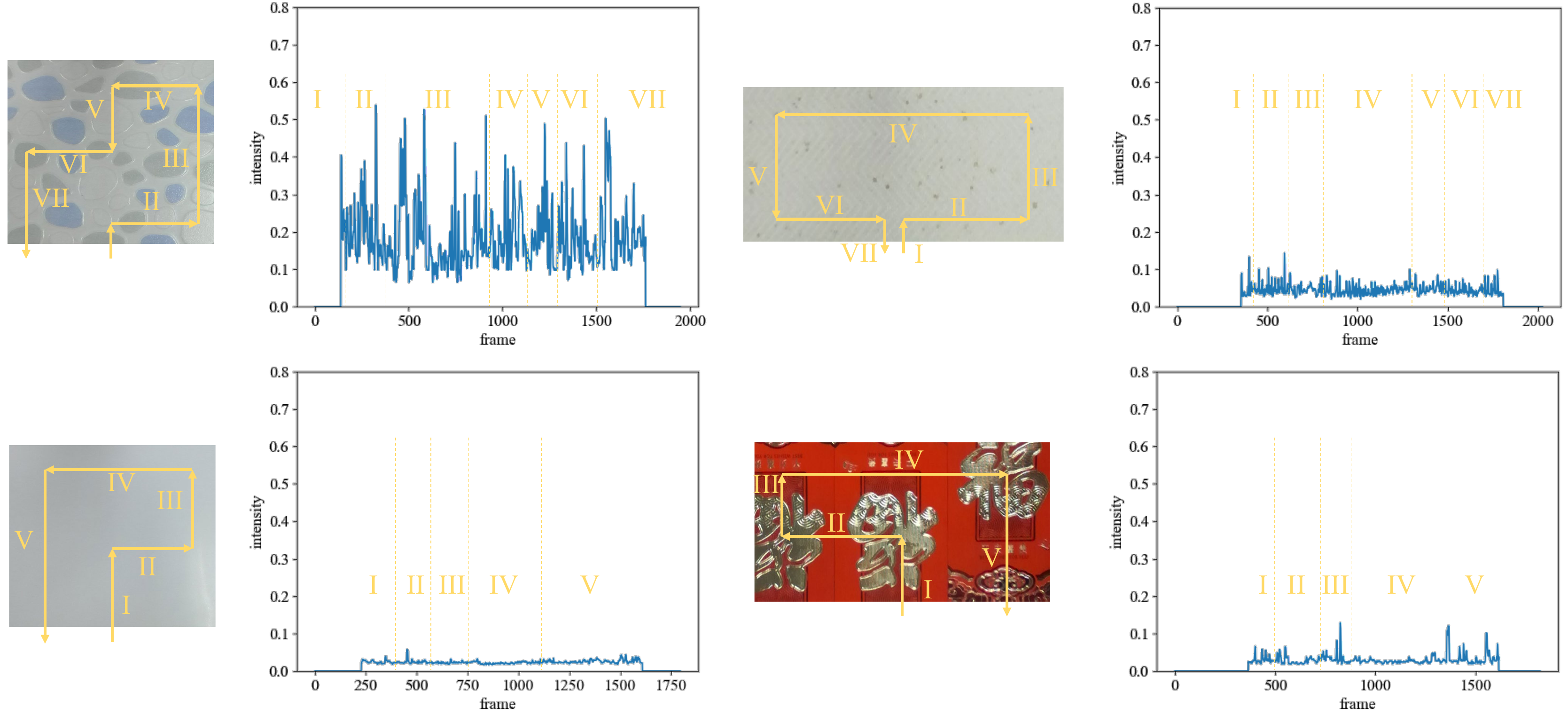}
\caption{\textbf{An illustration of joint visual and tactile perception with spatiotempral consistency.}}
\label{fig:app_result}
\end{figure*}

\section{V-Touching: An Exemplar VR Application} \label{sec:application}

To demonstrate the effectiveness of the developed scheme and the corresponding generated dataset, we build an interactive game in VR called \textbf{V-Touching}, as shown in~\cref{fig:application}.

We utilize a haptic glove as the input device for virtual hand poses, as well as the output device for haptic feedback. Let’s take a close look at the system. We assume a client-server architecture, which is currently implemented by using Socket connecting two local hosts in one computer. First, the poses, e.g. position, velocity signals of the virtual hand in the client side, transmits to the server. There is a mirror virtual hand in the server side that follows the movements to touch the surface of different materials. At the same time, the contact positions and contact material types are recorded. The system will present different friction, or we say, the amplitude of the vibrotactile signal, based on the contact position and contact depth. The tactile signal is rendered from real-world captured texture data. After we obtain the tactile signal, we push it to the tactile encoder, then transmit the encoded data to the client side, and finally reconstruct the signal via the decoder. The reconstructed data are then rendered and displayed via voice-coil like actuators in the tactile glove. We also assume that a virtual camera is used to capture the operation at the same time, which is transmitted to the client side. We utilize the data collected by the developed acquisition system in this application. The tactile data source is stored in a matrix format, which can be converted and stored as an image-based vibration map in the application. Each taxel of the vibration map corresponds to a specific tactile value based on its position. In particular, the virtual hand and textured objects are equipped with bounding boxes in the virtual environment. When the virtual hand makes contact with the textured surface, collision detection and raycasting techniques are employed to determine the contact position in world coordinates and visual texture coordinates (i.e. UV coordinates). By performing bilinear interpolation on the visual UV coordinates, we can retrieve the corresponding pixel from the texture map and align with the vibrotactile signal at that position.

For the implementation of \textbf{V-Touching}, we use Unity as the 3D rendering engine and the Avatar VR glove as the haptic input-output device. Since the Avatar VR glove provides an SDK(Software Development Kit) and the virtual hand model within the Unity engine, it allows us to directly control the poses of individual joints of the virtual hand. The gross pose of the virtual hand can be controlled using the HTC VIVE Tracker for accurate motion tracking. We also use keyboard for precise control of the horizontal movement of the virtual hand in the experiment.

In~\cref{fig:app_result}, we present four pairs of touch position variations along with the corresponding vibration waveforms of the \textbf{V-Touching} application. In the left image of each pair, the yellow arrows represent the moving trajectory of the user on the object surface, while the right figure illustrates the resulting vibrotactile intensity. To provide a clearer representation of the experimental outcomes, we have set the vertical axis range from 0 to 0.8. 

From~\cref{fig:app_result}, we can observe distinct vibration patterns for different objects, and regions with homogeneous and heterogeneous textures. This finding highlights the effectiveness of our application in providing users with spatiotemporal consistent experiences of visual and tactile modalities.

\section{Conclusions}\label{sec:conclusion}

In this paper, we developed a new data acquisition system for spatial-temporal consistent data collection of visual and vibrotactile signals. We also designed a tactile interaction application called \textbf{V-Touching} by using the dataset generated via the proposed system. Experimental analysis revealed that the data collected using the proposed system exhibits distinct vibration characteristics for different objects and regions. One limitation of this research is that we only allow the robotic arm to move in the Y direction during the data collection process. We should refine the data collection by incorporating sampling from multiple directions in order to enhance the accuracy of tactile information. We leave this for future research.


%



\bibliographystyle{IEEEtran}
\balance
\bibliography{reference}

\end{document}